# Brain Tumor Detection Based On Mathematical Analysis and Symmetry Information

Narkhede Sachin G., Prof. Vaishali Khairnar, Prof. Sujata Kadu

***Abstract***
Image segmentation some of the challenging issues on brain magnetic resonance (MR) image tumor segmentation caused by the weak correlation between magnetic resonance imaging (MRI) intensity and anatomical meaning. With the objective of utilizing more meaningful information to improve brain tumor segmentation, an approach which employs bilateral symmetry information as an additional feature for segmentation is proposed. This is motivated by potential performance improvement in the general automatic brain tumor segmentation systems which are important for many medical and scientific applications.
Brain Magnetic Resonance Imaging (MRI) segmentation is a complex problem in the field of medical imaging despite various presented methods. MR image of human brain can be divided into several sub-regions especially soft tissues such as gray matter, white matter and cerebrospinal fluid. Although edge information is the main clue in image segmentation, it can't get a better result in analysis the content of images without combining other information. Our goal is to detect the position and boundary of tumors automatically. Experiments were conducted on real pictures, and the results show that the algorithm is flexible and convenient.

## I. Introduction

Image segmentation is one of the primary steps in image analysis for object identification. The main aim is to recognize homogeneous regions within an image as distinct and belonging to different objects. Segmentation stage does not worry about the identity of the objects. They can be labeled later. The segmentation process can be based on finding the maximum homogeneity in grey levels within the regions identified [2].

Segmentation subdivides an image into its regions of components or objects and an important tool in medical image processing. As an initial step segmentation can be used for visualization and compression. Through identifying all pixels (for two dimensional image) or voxels (for three dimensional image) belonging to an object, segmentation of that particular object is achieved. In medical imaging, segmentation is vital for feature extraction, image measurements and image display [3].

As the first step in image analysis and pattern recognition, image segmentation is always a crucial component of most image analysis and pattern recognition systems, and determines the quality of the final result of analysis. So image segmentation has been intensively and extensively studied in the past years. And a wide variety of methods and algorithms are available to deal with the problem of segmentation of images. According to existing automatic image segmentation techniques can be classified into four categories, namely, (1) Clustering Methods, (2) Thresholding Methods, (3) Edge-Detection Methods, and (4) Region-Based Methods [6].

## II. Literature Survey

### 2.1 Magnetic Resonance Imaging (MRI)

Magnetic resonance imaging (MRI) is an imaging technique based on the physical phenomenon of Nuclear Magnetic Resonance (NMR). It is used in medical settings to produce images of the inside of the human body. MRI can produce an image of the NMR signal in a thin slice through the human body. By scanning a set of such slices a volume of a part of the human body can be represented with MRI.

### 2.2 Brain MR Images

MRI is an advanced medical imaging technique providing rich information about the human soft tissue anatomy. It has several advantages over other imaging techniques enabling it to provide 3-dimensional data with high contrast between soft tissues. However, the amount of data is far too much for manual analysis/interpretation, and this has been one of the biggest obstacles in the effective use of MRI. For this reason, automatic or semi-automatic techniques of computer-aided image analysis are necessary. Segmentation of MR images into different tissue classes, especially gray matter (GM), white matter (WM) and cerebrospinal fluid (CSF), is an important task. Brain MR images have a number of features, especially the following: Firstly, they are statistically simple; MR Images are theoretically piecewise constant with a small number of classes.

Secondly, they have relatively high contrast between different tissues. The contrast in an MR image depends upon the way the image is acquired. By altering radio frequency and gradient pulses and by carefully choosing relaxation timing, it is possible to highlight different component in the object being imaged and produce high contrast images. These two features facilitate segmentation [12].

**2.3 Digital Image Processing**

An image may be defined as a two-dimensional function f(x, y), where x & y are spatial coordinates, & the amplitude of f at any pair of coordinates (x, y) is called the intensity or gray level of the image at that point. Digital image is composed of a finite number of elements, each of which has a particular location & value. The elements are called pixels.

There are no clear-cut boundaries in the continuum from image processing at one end to complete vision at the other. However, one useful paradigm is to consider three types of computerized processes in this continuum: low-, mid-, & high-level processes. Low-level process involves primitive operations such as image processing to reduce noise, contrast enhancement & image sharpening. A low-level process is characterized by the fact that both its inputs & outputs are images. Mid-level process on images involves tasks such as segmentation, description of that object to reduce them to a form suitable for computer processing & classification of individual objects. A mid-level process is characterized by the fact that its inputs generally are images but its outputs are attributes extracted from those images. Finally higher- level processing involves "Making sense" of an ensemble of recognized objects, as in image analysis & at the far end of the continuum performing the cognitive functions normally associated with human vision. Digital image processing, as already defined is used successfully in a broad range of areas of exceptional social & economic value.

**2.4 Image Segmentation**

**2.4.1 What is Image Segmentation?**

Segmentation is the process of splitting an observed image into its homogeneous or constituent regions. The goal of segmentation is to simplify or change the representation of an image into something that is more meaningful and easier to analyze. It is important in many computer vision and image processing application. Image segmentation is an essential but critical component in low level vision, image analysis, pattern recognition, and now in robotic systems. Besides, it is one of the most difficult and challenging tasks in image processing, and determines the quality of the final results of the image analysis. Intuitively, image segmentation is the process of dividing an image into different regions such that each region is homogeneous while not the union of any two adjacent regions. An additional requirement would be that these regions had a correspondence to real homogeneous regions belonging to objects in the scene.

**III. Problem Statement**

Image segmentation is a key step from the image processing to image analysis, it occupy an important place. On the other hand, as the image segmentation, the target expression based on segmentation, the feature extraction and parameter measurement that converts the original image to more abstract and more compact form, it is possible to make high-level image analysis and understanding.

If the input brain image is colorized , it is converted into gray image. First read the red, blue and green value of each pixel and then after formulation, three different values are converted into gray value. The automated edge detection technique is proposed to detect the edges of the regions of interest on the digital images automatically. The method is employed to segment an image into two symmetric regions based on finding pixels that are of similar in nature. The more symmetrical the two regions have, the more the edges are weakened. At the same time, the edges not symmetrical are enhanced. In the end, according to the enhancing effect, the unsymmetrical regions can be detected, which is caused by brain tumor.

**IV. The Proposed Mechanism**

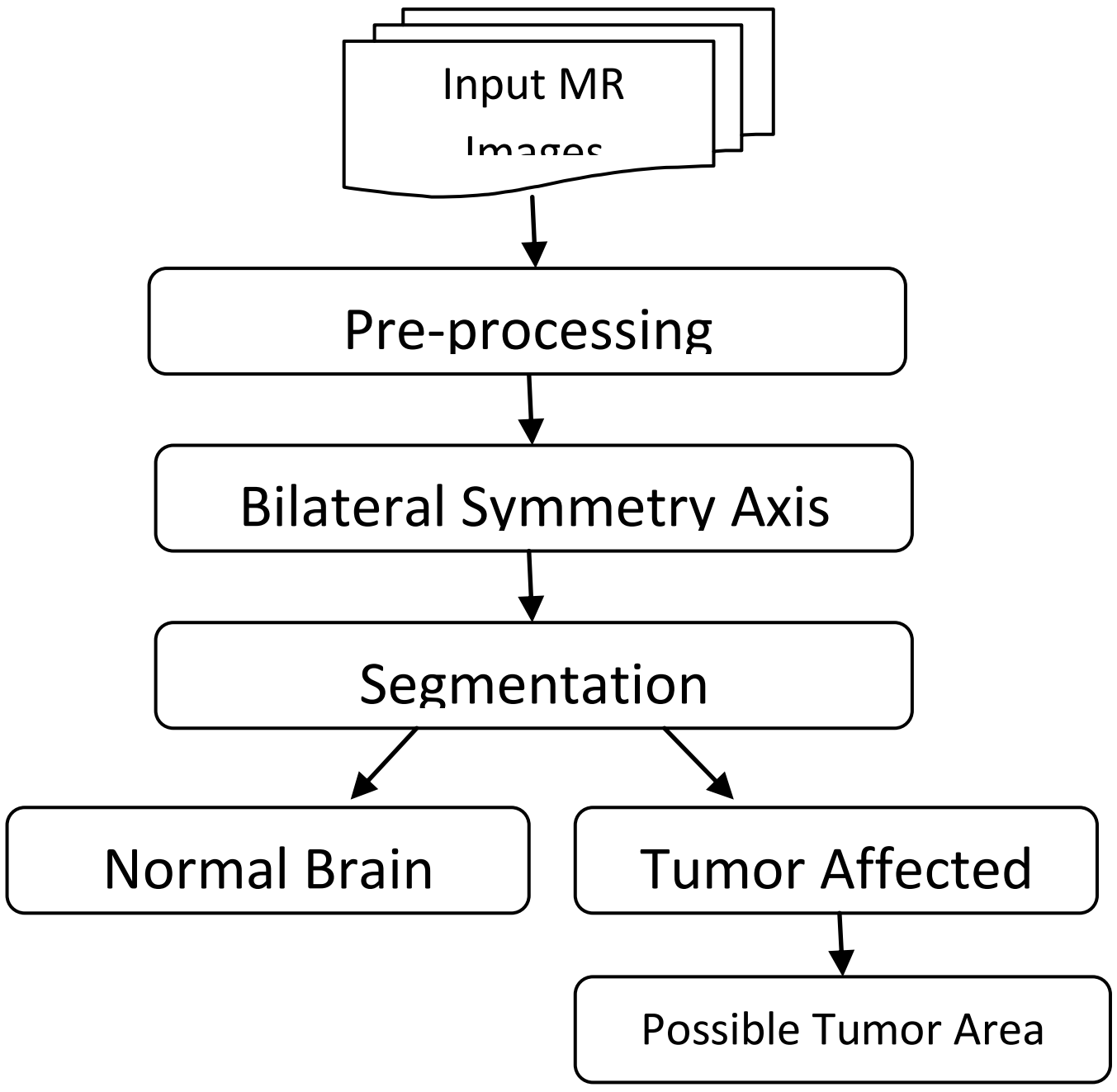

Figure 4.1: Proposed Model

## V. Methodology Used

Step 1:Use canny edge detection technique for the finding the edges in brain image.
Step 2:Determine the Mid pixel position of the row and read the intensity of Mid pixel of row
Input: Digitized brain image stored in a two dimensional array

- Read the next Row of image (initially the first row) to a single dimensional array i.e. Row
- Determine the Mid Pixel position of the Row and Read the Intensity of Mid Pixel of Row
- Check the Mid Pixel Intensity with all the pixels stating from First to Last position of Row
- If any Pixel Intensity of the Row is greater than Maximum Threshold (MT) then divide the Row into two equal parts. If the total positions of Row are odd, then take floor value of Middle position assign to Mid
- Push (First and Mid ) and (Mid+1 and Last) to stack respectively
- If any Pixel Intensity of the Row is not greater than MT then intensity of Mid pixel will be moved to all the pixels after modifying intensity value using uniform colour quantization technique in colour space breaking in eight level scales.
- If the stack is not empty then pop first and last position of Row and go to Step 2
- If it is not Last Row go to step 1

Step 3:Fit the curve by using LSM and Crammer rule.
Step 4:Show the curve in tumor affected area.
Step 5:Calculate and Show the tumor area by using Automatic brain tumor detection

**Symmetry axis defining:**

**Least Square Method:**

The method of least squares is a standard approach to the approximate solution of over determined systems, i.e., sets of equations in which there are more equations than unknowns. "Least squares" means that the overall solution minimizes the sum of the squares of the errors made in solving every single equation.

Use the least square method to get the symmetry curve line *y* approximatively

Expression for the straight line is

$$y= a_0 + a_1x + e$$

where $a_0$ and $a_1$ are coefficients representing the intercept & slope respectively

*e* is the error, or residual, between the model and the observations

$$e=y- a_0 - a_1x$$

Residual is the inconsistency between the true value of y and the approximate value, $a_0 + a_1x$ , predicted by the linear equation

*y* might be a linear function of $x_1$ and $x_2$

$$y= a_0 + a_1x1 + a_2x2+e$$

Such an equation is particularly useful when fitting experimental data where the variable being studied is often a function of two other variables. For this two-dimensional case the regression "line" becomes a "plane".

$$Sr=\sum_{i=1}^{n}(yi - a0 - a1x1.i - a2x2.i)^2$$

and differentiating with respect to each of the unknown coefficients:

$$\frac{\delta s_r}{\delta a_0} = -2\sum(yi - a0 - a1x1.i - a2x2.i)$$

$$\frac{\delta s_r}{\delta a_1} = -2\sum x_1.i.i(yi - a0 - a1x1.i - a2x2.i)$$

$$\frac{\delta s_r}{\delta a_2} = -2\sum x_2.i(yi - a0 - a1x1.i - a2x2.i)$$

The coefficients yielding the minimum sum of the squares of the residuals are obtained by setting the partial derivatives equal to zero and expressing the result in matrix form.

$$\begin{bmatrix} n & \sum x_1.i & \sum x_2.i \\ \sum x_1.i & \sum x_1^2.i & \sum x_1.ix_2.i \\ \sum x_2.i0 & \sum x_1.ix_2.i & \sum x_{21}^2.i \end{bmatrix} \begin{Bmatrix} a_0 \\ a_1 \\ a_2 \end{Bmatrix} = \begin{Bmatrix} \sum y_i \\ \sum x_1.iy_i \\ \sum x_2.iy_i \end{Bmatrix}$$

**Cramer's Rule**

Cramer's rule is another solution technique that is best suited to small numbers of equations. Before describing this method, we will briefly review the concept of the determinant, which is used to implement Cramer's rule. In addition, the determinant has relevance to the evaluation of the ill-conditioning of a matrix.

This rule states that each unknown in a system of linear algebraic equations may be expressed as a fraction of two determinants with denominator D and with the numerator obtained from D by replacing the column of coefficients of the unknown in question by the constant.

Table 5.1: Number of detected edges

| | | **Robert** | **Prewitt** | **Canny** |
|---|---|---|---|---|
| 1 | High | 5259 | 4382 | 1997 |
| 2 | High | 5120 | 4323 | 1836 |
| 3 | High | 6807 | 5757 | 2302 |
| 4 | Low | 1491 | 649 | 317 |
| 5 | Low | 2509 | 1080 | 433 |
| 6 | Low | 2567 | 1072 | 417 |

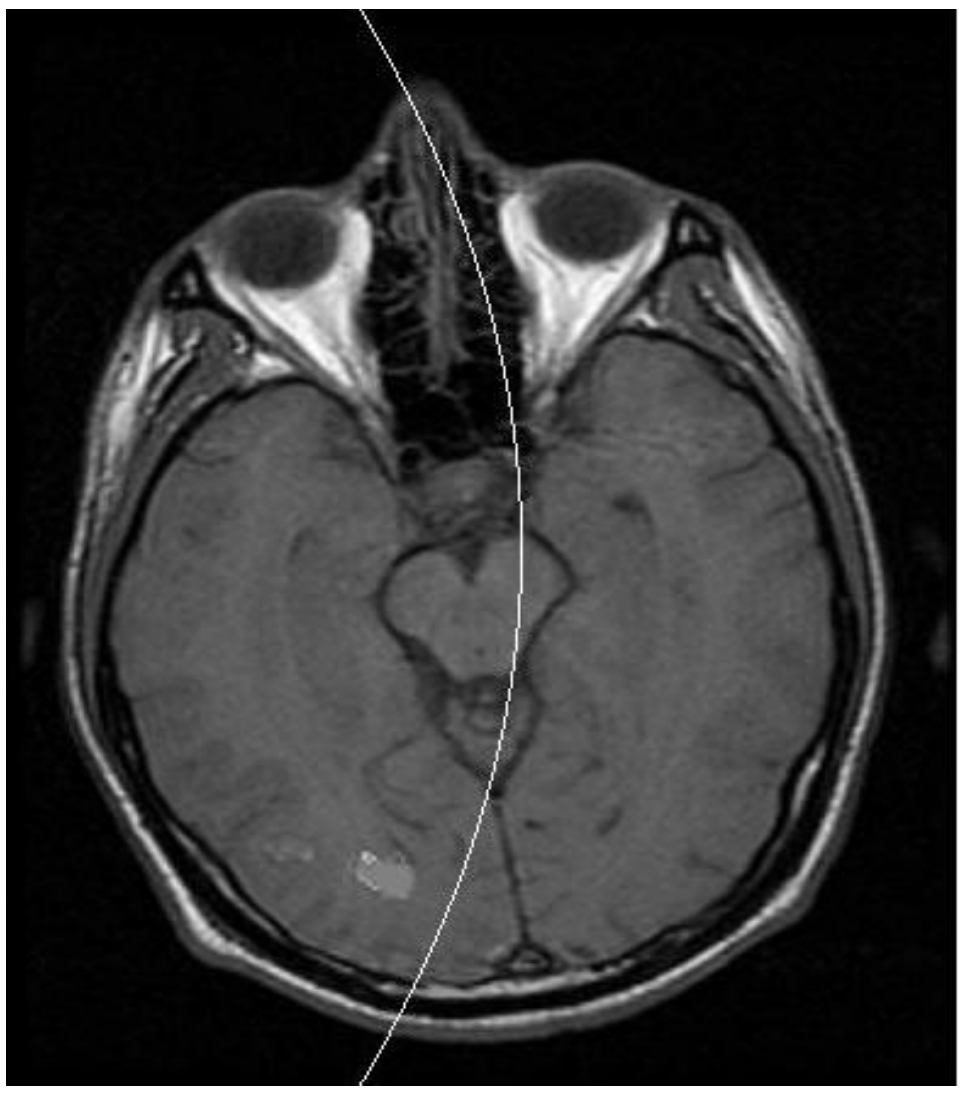

(a)High Grade

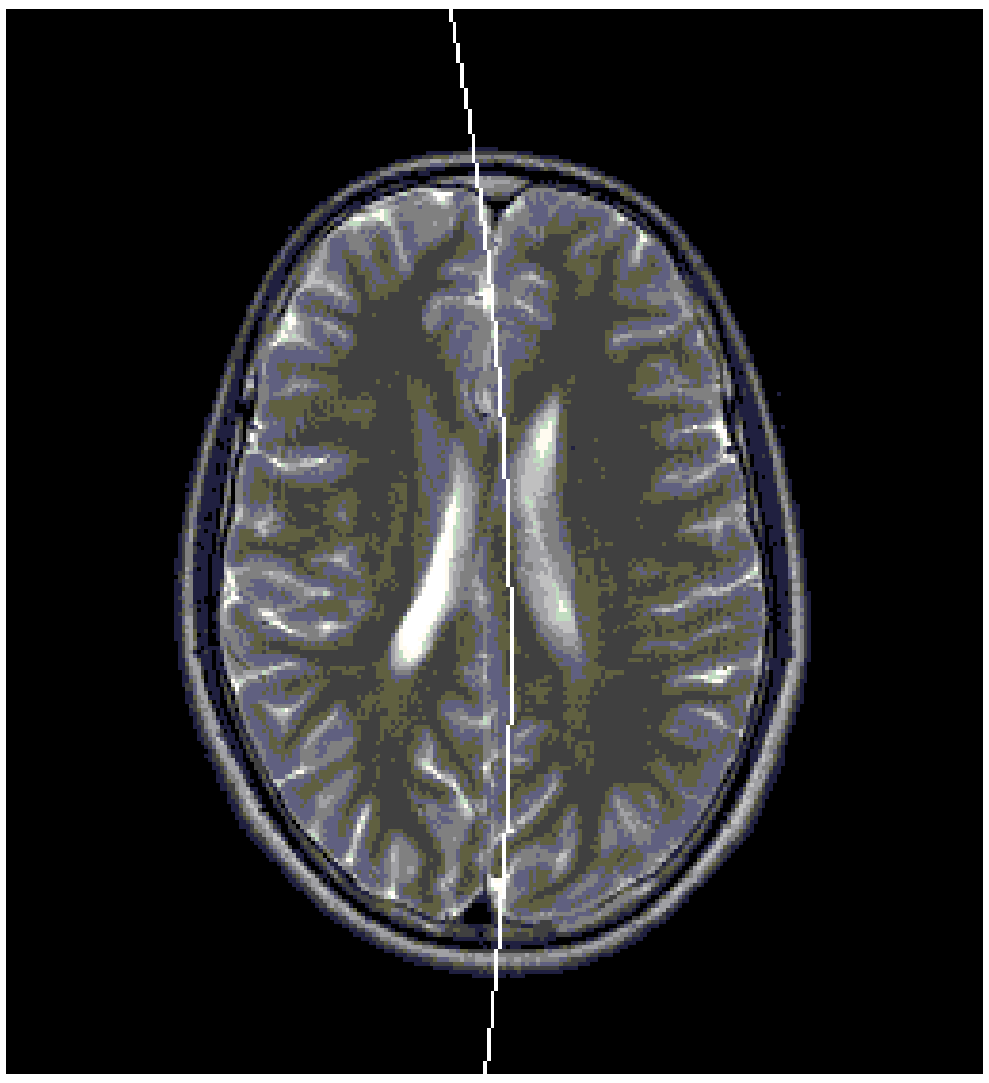

(b) Low Grade

Figure 5.1: Bilateral Axis

## VI. Conclusion

A new system that can be used as a second decision for the surgeons and radiologists is proposed.High grade tumor have more true edges than low grade.MRI of healthy brain has an obviously character almost bilateral symmetrical .However, if there is macroscopic tumor, the symmetry characteristic will be weakened According to the influence on the symmetry by the tumor, we develop a segment algorithm to detect the tumor region automatically